\documentclass[10pt,twocolumn,letterpaper]{article}

\usepackage{iccv}
\usepackage{times}
\usepackage{epsfig}
\usepackage{graphicx}
\usepackage{amsmath}
\usepackage{amssymb}
\usepackage{bm}

\usepackage[pagebackref=true,breaklinks=true,letterpaper=true,colorlinks,bookmarks=false]{hyperref}

 \iccvfinalcopy 

\def\iccvPaperID{****}

\ificcvfinal\fi
\begin{document}

\title{iqiyi Submission to ActivityNet Challenge 2019 Kinetics-700 challenge: Hierarchical Group-wise Attention}

\author{Qian Liu, Dongyang Cai, Jie Liu, Nan Ding, Tao Wang\\
iqiyi, Inc.\\ 
{\tt\small }
}

\maketitle

\begin{abstract}
	
	In this report, the method for the iqiyi
	submission to the task of ActivityNet 2019 Kinetics-700 challenge is described.
	Three models are involved in the model ensemble stage: TSN, HG-NL and StNet.
	We propose the hierarchical group-wise non-local (HG-NL) module for frame-level features aggregation for video classification. The standard non-local (NL) module is effective in aggregating frame-level features on the task of video classification but presents low parameters efficiency and high computational cost.
	The HG-NL method involves a hierarchical group-wise structure and generates multiple attention maps to enhance performance. Basing on this hierarchical group-wise structure, the proposed method has competitive accuracy, fewer parameters and smaller computational cost than the standard NL.
	For the task of ActivityNet 2019 Kinetics-700 challenge, after model ensemble, we finally obtain an averaged top-1 and top-5 error percentage 28.444\% on the test set.
  
\end{abstract}

\section{Introduction}

\begin{figure*}
	\begin{center}
		\includegraphics[width=0.75\linewidth]{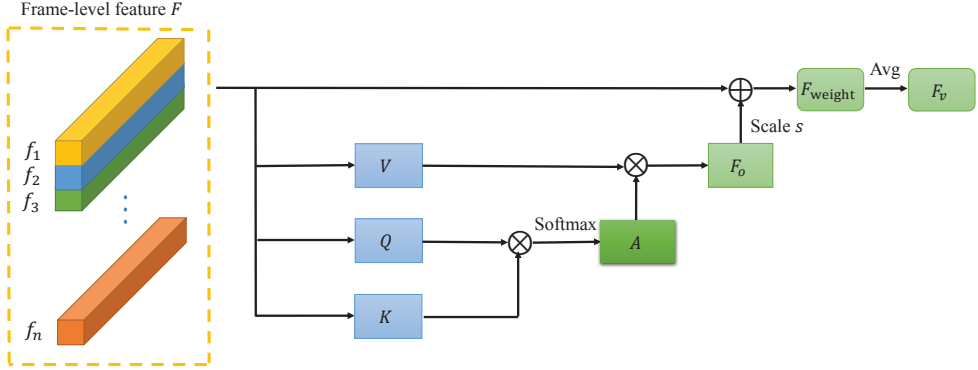}
	\end{center}
	\caption{Self-Attention (Non-local) Based Frame-level Features Aggregation ($\oplus$ denotes element-wise sum and $\otimes$ denotes matrix multiplication).}
	\label{Figure3-2}
\end{figure*}

\begin{table*}
	\begin{center}
		\caption{The number of parameters in NL and HG-NL ($m = 1024, g_1 = 16$ and $g_2 = 8$). The parameters of HG-NL is about 8 - 14 times fewer than NL method. It is roughly 70 times fewer if parameters are shared across groups in a grouped convolutional layer in HG-NL. }
		\label{Tabel-3-1}
		\begin{tabular}{|l|c|c|c|c|}
			\hline
			&NL ($m_1 = m/2$) & NL ($m_1 = m/8$) & HG-NL ($m_1 = m/8$)& HG-NL (shared parameters, $m_1 = m/8$) \\
			\hline
			$W_q$  &  0.5248M   &  0.1312M  & 8.32K &0.52k\\
			\hline
			$W_k$  &  0.5248M  &  0. 1312M & 8.32K  &0.52k\\
			\hline
			$W_v$   &  1.0496M   &  1.0496M     &  132.096k &16.512k\\
			\hline
			Others  &  --	 & 	-- 	 & 	--& 	--\\
			\hline
			All  & 2.0992M  &  1.312M	 & 148.736K&17.552k\\
			\hline
		\end{tabular}
	\end{center}
\end{table*}

\begin{table*}
	\begin{center}
		\caption{MAdds (multiply-adds) of NL and HG-NL. Each convolution layer in HG-NL has g1(=16) or g2(=8) times fewer MAdds than NL. The MAdds of other non-convolution layers keep roughly unchanged.}
		\label{Tabel-3-2}
		\begin{tabular}{|l|c|c|c|}
			\hline
			&NL ($m_1 = m/2$) & NL ($m_1 = m/8$) & HG-NL ($m_1 = m/8$) \\
			\hline
			$W_qF$  & $m^2n$  &  $m^2n/4$   & $m^2n/(4g_1)$ \\
			\hline
			$W_kF$    &  $m^2n$  &  $m^2n/4$  & $m^2n/(4g_1)$ \\
			\hline
			$W_vF$   &   $2m^2n$    &  $2m^2n$     & $2m^2n/g_2$ \\
			\hline
			$Q^TK / Gmm(Q,K)$  & $n^2m-n^2$  &  $n^2m/4-n^2$  &  $n^2m/4-g_2n^2$ \\
			\hline
			$\mathrm{Softmax}(\cdot)$ / $\mathrm{Relu}(\cdot)$   & --   &   --     &     -- \\
			\hline
			$VA / Gmm(V, A)$   &  $mn(2n-1) $      &      $mn(2n-1) $     &      $ mn(2n-1)$ \\
			\hline
		\end{tabular}
	\end{center}
\end{table*}

Video classification is one of the challenging tasks in computer vision.
Publicly challenges and available video datasets accelerate the research processing,
especially the ActivityNet series challenges and related datasets.
In recent years, deep convolutional neural networks (CNNs) bring remarkable improvements on the accuracy of video classification \cite{wang2016temporal, carreira2017quo, qiu2017learning, feichtenhofer2018slowfast}. 

In this report, the method for the iqiyi
submission to the trimmed activity recognition
(Kinetics) tasks of the ActivityNet Large Scale Activity
Recognition Challenge 2019 is described.
The Kinetics-700 dataset covers 700 human action classes and consists of approximately 650,000 video clips. And, each clip lasts around 10 seconds.

In our model ensemble stage, three models are involved: TSN\cite{wang2016temporal}, HG-NL and StNet\cite{he2018stnet}.
We propose the hierarchical group-wise non-local (HG-NL) module for frame-level features aggregation for video classification. 

Frequently-used aggregating methods include maximum, evenly averaging and weighted averaging. 
The NL module in \cite{wang2018non} is also able to be used for aggregating frame-level features.
However, the NL module in \cite{wang2018non} presents low parameters efficiency and high computational cost, as discussed in detail later in this paper. 

We address the problem of building a highly efficient self-attention based frame-level features aggregation module.
The Hierarchical Group-wise Non-Local (HG-NL) module for frame-level features aggregation is proposed.
Comparison with NL in \cite{wang2018non}, the HG-NL module has fewer parameters and smaller computational cost.
The proposed module involves a hierarchical group-wise structure, which includes the primary grouped convolutions and the secondary grouped matrix multiplication. 
Moreover, HG-NL generates multiple attention maps. It brings one attention map for each feature group in the entire feature matrix and can mine the non-local information in features in detail.

\section{Method}

\subsection{HG-NL}
In this section, the HG-NL is presented in detail. 

\subsubsection{Formulation of Frame-level Features Aggregation}
Considering a video $v$, a sequence of frames $ \{s_1, s_2, … , s_n\}$ ($n$ is the length of a sequence of frames) are extracted from the entire video via some specific rules. 

The feature information of a single frame is obtained via a pre-trained convolution network:
\begin{equation}\label{3-1-1}f_i = C(s_i),\end{equation}  
where $s_i$ denotes the $i$-th frame,  $f_i$ is the feature information of $s_i$ ,  and $C(\cdot)$ denotes the ConvNet operating.  

The compact video-level features can be obtained via aggregating the features from multiple frames:
\begin{equation}
	\label{3-1-2}
	F_v = Agg(f_1,f_2,…,f_n),
\end{equation}
where $Agg(\cdot)$ is the aggregating function, $n$ is the length of a sequence of frames, and $F_v$ denotes the compact video-level features. 

\begin{figure*}
	\begin{center}
		\includegraphics[width=0.75\linewidth]{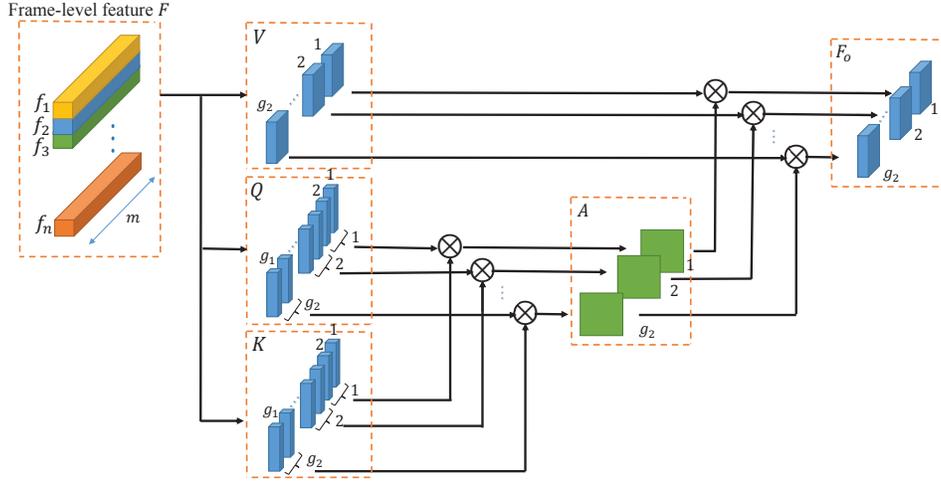}
	\end{center}
	\caption{Hierarchical Group-wise Non-local module ($g_1 = 2g_2$). $V$, $Q$ and $K$ are obtained via grouped convolutions. $A$ and $F_o$ are obtained using grouped matrix multiplication.}
	\label{Figure3-3}
\end{figure*}

\subsubsection{Self-Attention (Non-local) Based Frame-level Features Aggregation}
\label{Sec:Self-Attention}

In self-attention module, the response of a position is computed with weighted average of all positions in an embedding space. 
As a representative module of attention mechanism, the NL in \cite{wang2018non} is adopted to aggregate frame-level features here and is able to obtain long-rang dependencies across the frames.

Let $F'=[f_1, f_2, …, f_n]\in \mathbb{R}^{m\times n}$, where $m$ is the length of each frame's feature vector.
$F\in \mathbb{R}^{m\times n \times1}$, which denotes the feature information of $n$ frames, can be obtained via reshaping the size of $F'$ to $m\times n \times1$ (corresponding to $C*H*W$).
Then, an attention map having the size of $n\times n$ and containing the relationships between every pair of frames can be obtained 
\begin{equation}\label{3-2-2}
	A = \mathrm{softmax}(Q^TK), 
\end{equation}
where $Q=W_qF$, $K=W_kF$, and weight matrices $W_q\in\mathbb{R}^{m_1\times m}$ and $W_k\in\mathbb{R}^{m_1\times m}$ are learned parameters.
Commonly, weight matrices $W_q$, $W_k$ are implemented as $1 \times 1$ convolutions. 

The output based on the attention map $A$ is
\begin{equation}F_o = VA,\end{equation}
where $V=W_vF$ and weight matrices $W_v\in\mathbb{R}^{m\times m}$ is also operated as $1\times1$ convolutions. 

After this, $F_{\mathrm{weight}}$ can be obtained
\begin{equation}\label{3-2-3}F_{\mathrm{weight}} = s \cdot F_o + F.\end{equation}
In the above formulation, $s$ is a scale parameter and the output $F_{\mathrm{weight}}$ has the same size as the input signal $F$.

The video-level feature $F_v$ is obtained via evenly averaging of $F_{\mathrm{weight}}$
\begin{equation}\label{3-2-4}F_v = \mathrm{avg}(F_{\mathrm{weight}}).\end{equation}

Figure \ref{Figure3-2} shows the schema of the NL module for frame-level features aggregation.

\textbf{Analysis of NL module} The NL module is effective for aggregating frame-level features.
However, the NL module presents low parameters efficiency and high computational cost.
The number of parameters in the NL module is computed as follows. 
For convolution layers corresponding to $ W_q $, $ W_k $ and $ W_v $, the number of their parameters is $ (1 \times 1 \times m \times m_1 + m_1)$, $(1 \times 1 \times m \times m_1 + m_1)$ and $(1 \times 1 \times m \times m + m)$ individually. 
When $m = 1024$, the number of parameters in the NL module can be computed and shown in Table \ref{Tabel-3-1}. 
As shown in Table \ref{Tabel-3-1}, if $m_1 = m/8$, the number of parameters is about 1.31M. If $m_1 = m/2$, the number of parameters is about 2M. The number is quite large for the practical use. 
In contrast, many backbone networks have very small number of parameters, such as MobileNetV2 \cite{sandler2018mobilenetv2} (3.4M), MobileNetV2-1.4 (6.9M), MF-Net-2D (5.8M), MF-Net-3D \cite{Chen_2018_ECCV} (8.0M) and I3D-RGB \cite{carreira2017quo} (12.1 M).                   
As for the computational complexity, when $m = 1024$, the total number of multiply-adds (MAdds) required in convolution layers in NL is about $4n$ M (when $m_1 = m/2$) and $2.5n$ M (when $m_1 = m/8$).
Therefore, it makes sense to reduces parameters redundancy and computational cost of NL module.

\subsubsection{Hierarchical Group-wise Non-local Module}

In order to reduce parameters redundancy and computational cost, the Hierarchical Group-wise Non-local (HG-NL) module for fame-level features aggregation is proposed. HG-NL has the hierarchical group-wise structure and generates several attention maps.
The HG-NL module for fame-level features aggregation is performed as following. 

Firstly, in HG-NL, weight matrices $W_q$, $W_k$ are implemented as $1 \times 1$ grouped convolutions with the number of groups being $g_1$. The grouped convolutions can reduce the parameters and the number of operations measured by MAdds largely. 

After this, the attention map $A$ is computed:
\begin{equation}\label{3-3-1}A = \mathrm{Relu}(\mathrm{Gmm}(Q,K)), \end{equation}
where $\mathrm{Gmm}(\cdot)$ denotes the grouped matrix multiplication with the number of groups being $g_2$, $A \in \mathbb{R}^{g_2 \times n \times n}$ includes $g_2$ attention maps, and each attention map has size $n \times n$.

As shown in Figure \ref{Figure3-3}, the grouped matrix multiplication in Eq \eqref{3-3-1} brings one attention map for each feature group in $V$, and the number of attention map achieves $g_2$. 
This can mine the non-local information in features more detailedly and effectively. As for the NL, only one attention map occurs.
Besides, the softmax is deleted in HG-NL. 
The computation of $\mathrm{Relu}(\cdot)$ in Eq \eqref{3-3-1} is lightweight, and the $\mathrm{Relu}(\cdot)$ can provide the non-linearity for the HG-NL module. 

Then, keeping the same groups as in the grouped matrix multiplication in Eq \eqref{3-3-1}, weight matrices $W_v$ is operated as $1\times1$ grouped convolutions with the number of groups being $g_2$ and $F_o$ is computed via the grouped matrix multiplication with the number of groups being $g_2$
\begin{equation}F_o = \mathrm{Gmm}(V, A).\end{equation}

At last, $F_v$ can be obtained based on $F_o$ via Eq \eqref{3-2-3} and Eq \eqref{3-2-4} in Section \ref{Sec:Self-Attention}.

Figure \ref{Figure3-3} shows the schema of the HG-NL module ($g_1 = 2g_2$).  
In general, let $g_1 = rg_2$ and $r$ is a ratio. Then the relationship of $g_1$ (primary grouped convolutions) and $g_2$ (secondary grouped matrix multiplication) forms the hierarchical group-wise structure. 
Consider the value of $g_1$ and $g_2$. 
Even though multiple attention maps are able to mine the non-local information more detailedly, each attention map will cover too narrow feature information if $g_2$ is too big. On the other hand, when $g_1$ is bigger, the related parameters and MAdds is smaller. 
Therefore, in common, the values of $g_1$ and $g_2$ are set to different values. As a special case, when $g_1$ equals $g_2$, the effect of HG-NL is the same as processing each feature group of $F$ via NL module individually. 

\textbf{Analysis of HG-NL module} For convolution layers corresponding to $ W_q $, $ W_k $ and $ W_v $, the number of parameters of them is $ g_1 \times ( 1 \times 1 \times (m/ g_1) \times (m_1/ g_1) + m_1/ g_1) $, $ g_1 \times ( 1 \times 1 \times (m/ g_1) \times (m_1/ g_1) + m_1/ g_1) $ and $ g_2 \times ( 1 \times 1 \times (m/ g_2) \times (m/ g_2) + m/g_2) $ individually. 
As shown in Table \ref{Tabel-3-1}, when $m = 1024, g_1 = 16$ and $g_2 = 8$, the HG-NL only requires about 1:8 - 1:14 times fewer parameters than the NL, which has roughly 1.31M ($m_1 = m/8$) - 2.1M ($m_1 = m/2$) parameters. If parameters are shared across groups in a grouped convolutional layer in HG-NL, the number of each convolution layer's parameters will be further reduced $g_1$ or $g_2$ times.
Besides, as shown in Table \ref{Tabel-3-2}, when $m = 1024$, the MAdds required in convolution layers in HG-NL is about $(0.5n/g_1 + 2n/g_2)$ M and is several times fewer MAdds than convolution layers in the NL. The MAdds required in other non-convolution layers keep roughly unchanged. 
Thus, as we can see that the HG-NL is able to reduce the model redundancy and computational cost.
Meanwhile, HG-NL can achieve the competitive accuracy as NL.

\subsubsection{Implementation of HG-NL for Video Classification}

Benefiting from that no fully-connected layers are included in the network architecture of HG-NL, $n$ (the number of frames) can be arbitrarily adjusted. Thus, in the evaluation phase of the proposed HG-NL module, the number of frames selected from a video for predicting
the label is not needed fixed as the same value as in the training phase and can be adjusted. 

\begin{table*}
	\begin{center}
		\caption{Results of models on Kinetics-700 val set.}	
		\label{Tabel-4-1}
		\begin{tabular}{|l|c|c|}
			\hline
			Model & Val Accuracy in train phase (3 segments): Top-1 ($\%$)  & Val Accuracy in test phase (25 segments): Top-1 ($\%$)   \\
			\hline
			TSN   &  57.38 & 61.83\\ 		
			\hline
			HG-NL & 57.713 & 62.12  \\
			\hline
			StNet &  55.7  & - \\
			\hline
		\end{tabular}
	\end{center}
\end{table*}

\begin{table}
	\begin{center}
		\caption{Results on Kinetics-700 test set. The avg. error is an averaged top-1 and top-5 error. }
		\label{Tabel-4-2}
			\begin{tabular}{|l|c|}
		\hline
		Models &   avg.error
		 \\
		\hline
		Model Ensemble  &   0.28444 \\ 		
		\hline
	\end{tabular}
	\end{center}
\end{table}

\subsection{Model Ensemble}

In the model ensemble stage, three models are involved: TSN\cite{wang2016temporal}, HG-NL and StNet\cite{he2018stnet}.

\section{Experiments}
\label{Sec-4}

In this section, we report some experimental results on Kinetics-700 dataset of our method.
All models are pre-trained on the Kinetics-600 training set. We finetuned these models on the Kinetics-700 training set. Se-Resnext101 is adopted as the backbone network.
Due to the limited time, we exploit only RGB information.
In our experimens, the full-length video is divided into several equal segments, some frames are randomly selected from each segment.
During our training, the number of segments is set to 3 and one frame are randomly selected from each segment.
During evaluation, we follow the same testing setup as in TSN \cite{wang2016temporal}. 

\subsection{TSN}
\label{Sec-4-tsn}
In TSN experiments, the initial learning rate is set as 0.001 and decayed by a factor of 10 at 20 epochs and 30 epochs. The maximum iteration is set as 40 epochs.

\subsection{HG-NL}
In HG-NL experiments, $m_1 = \frac{m}{8}$, $g_1 = 16$, $g_2 = 8$.
Due to time limits, we finetuned the HG-NL on the Kinetics-700 training set for only 8 epochs with the model pre-trained by TSN in Section \ref{Sec-4-tsn}.
The initial learning rate is set as 0.001 and decayed by a factor of 10 at 4 epochs and 6 epochs. The maximum iteration is set as 8 epochs.
The results are shown in Table \ref{Tabel-4-1}.
We can see that HG-NL can obtain the top-1 accuracy of 62.12\%, compared with the top-1 accuracy of 61.83\% of TSN on the Kinetics-700 validation set, as shown in Table \ref{Tabel-4-1}.

\subsection{StNet}
For StNet\cite{he2018stnet}, the Temporal Modeling Block and Temporal Xception Block are used in our network. We adopt the same input of TSN as the input of StNet. 
Because of the time limits, we only trained the network for 20 epochs on kinetics-700 datasets.
The results of the StNet on kinetics-700 validation dataset is 55.7\% for top1 and 78.3\% for top 5 in the train phase(3-frames test).
\subsection{Model Ensemble}
\label{4.3.1-Measuring}

Three models are involved in the model ensemble stage: TSN\cite{wang2016temporal}, HG-NL and StNet\cite{he2018stnet}. Our team finally obtains an averaged top-1 and top-5 error percentage of 28.444\% on the Kinetics-700 test set.

\section{Conclusion}

In this report, our team’s solution to the task of ActivityNet 2019 Kinetics-700 challenge is described.
Experiment results have evidenced the effectiveness of the proposed HG-NL method. HG-NL achieves the better accuracy than the TSN baseline.
With the help of the hierarchical group-wise structure, the HG-NL module has 8 - 70 times fewer parameters and several times smaller computational complexity than the NL module.
After model ensemble, our team finally obtains an averaged top-1 and top-5 error percentage of 28.444\% on the Kinetics-700 test set.

{\small
\bibliographystyle{ieee}
\bibliography{K700-report-arxiv-202001}
}

\end{document}